# Bayesian Model Averaging Using the $k$-best Bayesian Network Structures

**Jin Tian, Ru He and Lavanya Ram**
Department of Computer Science
Iowa State University
Ames, IA 50011
{*jtian, rhe, lram*}@iastate.edu

## Abstract

We study the problem of learning Bayesian network structures from data. We develop an algorithm for finding the $k$-best Bayesian network structures. We propose to compute the posterior probabilities of hypotheses of interest by Bayesian model averaging over the $k$-best Bayesian networks. We present empirical results on structural discovery over several real and synthetic data sets and show that the method outperforms the model selection method and the state-of-the-art MCMC methods.

## 1 Introduction

Bayesian networks (BN) are being widely used in various data mining tasks for probabilistic inference and causal modeling [12, 15]. One major challenge in the applications of BN is to learn the structures of BNs from data. In the Bayesian approach, we provide a prior probability distribution over the space of possible Bayesian networks and then compute the posterior distributions $P(G|D)$ of the network structure $G$ given data $D$. We can then compute the posterior probability of any hypothesis of interest by averaging over all possible networks. In some applications we are interested in structural features. For example, in causal discovery, we are interested in the causal relations among variables, represented by the edges in the network structure [7]. In other applications we are interested in predicting the posterior probabilities of new observations, for example, in classification tasks.

The number of possible network structures is superexponential $O(n!2^{n(n-1)/2})$ in the number of variables $n$. For example, there are about $10^4$ directed acyclic graphs (DAGs) on 5 nodes, and $10^{18}$ DAGs on 10 nodes. As a result, it is impractical to sum over all possible structures unless for tiny domains (less than 7 variables). The most common solution is to use model selection approach in which we use the relative posterior probability $P(D, G)$ (or other measures) as a *scoring metric* and we attempt to find a single network with the best score, the MAP network. We then use that model (or its Markov equivalence class) to make future predictions. This may be a good approximation if the amount of data is large relative to the size of the model such that the posterior is sharply peaked around the MAP model. However, in domains where the amount of data is small relative to the size of the model there are often many high-scoring models with non-negligible posterior. In this situation using a single model could lead to unwarranted conclusions about the structure features and also poor predictions about new observations. For example, the edges that appear in the MAP model do not necessarily appear in other approximately equally likely models. Also the model selection may be sensitive to the data samples given in the sense that a different set of data (from the same distribution) might well lead to a different MAP model. In such cases, using Bayesian model averaging is preferred.

Recently there has been progress on computing exact posterior probabilities of structural features such as edges or subnetworks using dynamic programming techniques [8, 9, 16]. These techniques have exponential time and memory complexity and are capable of handling data sets with up to around 20 variables. One problem with these algorithms is that they can only computer posteriors over modular features such as directed edges but can not compute non-modular features such as paths ("is there a path between nodes $X$ and $Y$"). Another problem is that it is very expensive to perform data prediction tasks. They can compute the exact posterior of new observational data $P(x|D)$ but the algorithms have to be re-run for each new data case $x$.

When computing exact posterior probabilities of features is not feasible, one solution that has been proposed is to approximate full Bayesian model averaging by finding a set of high-scoring networks and making prediction using these models [7, 10]. This leaves open the question of how to construct the set of representative models. One possible approach is to use the bootstrap technique which has been studied in [4]. However there are still many questions that need further study on how to use the bootstrap to approximate the Bayesian posterior.

One theoretically well-founded approach is to use Markov Chain Monte Carlo (MCMC) techniques. Madigan and York [11] used MCMC algorithm in the space of network structures (i.e., DAGs). Friedman and Koller [5] developed a MCMC procedure in the space of node orderings which was shown to be more efficient than MCMC in the space of DAGs and to outperform the bootstrap approach in [4] as well. Eaton and Murphy [3] developed a hybrid MCMC method (DP+MCMC) that first uses the dynamic programming technique in [8] to develop a global proposal distribution and then runs MCMC in the DAG space. Their experiments showed that the DP+MCMC algorithm converged faster than previous two methods [5, 11] and resulted in more accurate structure learning. One common problem to the MCMC and bootstrap approach is that there is no guarantee on the quality of the approximation in finite runs.

Madigan and Raftery [10] has proposed to discard the models whose posterior probability is much lower than the best ones (as well as complex models whose posterior probability is lower than some simpler one). In this paper, we study the approach of approximating Bayesian model averaging using a set of best Bayesian networks. It is intuitive to make predictions using a set of the best models, and we believe it is due to the computational difficulties of actually finding the best networks that this idea has not been systematically studied. In this paper, we develop an algorithm for finding the $k$-best network structures by generalizing the dynamic programming algorithm for finding the optimal Bayesian network structures in [13, 14]. We demonstrate the algorithm on several real data sets from the UCI Machine Learning Repository [1] and synthetic data sets from a gold-standard network. We then empirically study the quality of Bayesian model averaging using the $k$-best networks in structure discovery and show that the method outperforms the model selection method and the state-of-the-art MCMC methods.

## 2 Bayesian Learning of Bayesian Networks

A Bayesian network is a DAG $G$ that encodes a joint probability distribution over a set $X = \{X_1, \ldots, X_n\}$ of random variables with each node of the graph representing a variable in $X$. For convenience we will typically work on the index set $V = \{1, \ldots, n\}$ and represent a variable $X_i$ by its index $i$. We use $X_{Pa_i} \subseteq X$ to represent the set of parents of $X_i$ in a DAG $G$ and use $Pa_i \subseteq V$ to represent the corresponding index set.

In the problem of learning BNs from data we are given a training data set $D = \{x^1, x^2, \ldots, x^N\}$, where each $x^i$ is a particular instantiation over the set of variables $X$. In this paper we only consider situations where the data are complete, that is, every variable in $X$ is assigned a value. In the Bayesian approach to learning Bayesian networks from the training data $D$, we compute the posterior probability of a network $G$ as

$$P(G|D) = \frac{P(D|G)P(G)}{P(D)}. \tag{1}$$

Assuming global/local parameter independence, and parameter/structure modularity, $\ln P(D|G)P(G)$ can be decomposed into a summation of so-called local scores as [2, 6]

$$\ln P(G, D) = \sum_{i=1}^{n} score_i(Pa_i : D) \equiv score(G : D), \tag{2}$$

where, with appropriate parameter priors, $score_i(Pa_i : D)$ has a closed form solution. In this paper we will focus on discrete random variables assuming that each variable $X_i$ can take values from a finite domain. We will use the popular BDe score for $score_i(Pa_i : D)$ and we refer to [6] for its detailed expression. Often for convenience we will omit mentioning $D$ explicitly and use $score_i(Pa_i)$ and $score(G)$.

In the Bayesian framework, we compute the posterior probability of any hypothesis of interest $h$ by averaging over all possible networks.

$$P(h|D) = \sum_G P(h|G, D)P(G|D). \tag{3}$$

Since the number of possible DAGs is superexponential in the number of variables $n$, it is impractical to sum over all DAGs unless for very small networks. One solution is to approximate this exhaustive enumeration by using a selected set of models in $\mathcal{G}$

$$\hat{P}(h|D) = \frac{\sum_{G \in \mathcal{G}} P(h|G, D)P(G|D)}{\sum_{G \in \mathcal{G}} P(G|D)} \tag{4}$$

$$= \frac{\sum_{G \in \mathcal{G}} P(h|G, D)P(G, D)}{\sum_{G \in \mathcal{G}} P(G, D)}, \tag{5}$$

where $\hat{P}(.)$ denote approximated probabilities. In the model selection approach, we find a high-scoring model $G_s$ and use it to make predictions:

$$\hat{P}(h|D) = P(h|G_s, D). \tag{6}$$

In this paper we will perform model averaging using the set $\mathcal{G}$ of $k$-best networks.

We can estimate the posterior probability of a network $G \in \mathcal{G}$ as

$$\hat{P}(G|D) = \frac{P(G, D)}{\sum_{G \in \mathcal{G}} P(G, D)}. \tag{7}$$

Since $\sum_{G \in \mathcal{G}} P(G, D)$ is not greater than $P(D)$, the estimate $\hat{P}(G|D)$ is always an upper bound of $P(G|D)$, and it is a good estimate only if $\sum_{G \in \mathcal{G}} P(G|D)$ is close to 1. We can then estimate the posterior probability of hypothesis $h$ by

$$\hat{P}(h|D) = \sum_{G \in \mathcal{G}} P(h|G, D)\hat{P}(G|D). \quad (8)$$

If we are interested in computing the posteriors of structural features such as edges, paths, Markov Blankets, etc.. Let $f$ be a structural feature represented by an indicator function such that $f(G)$ is 1 if the feature is present in $G$ and 0 otherwise. We have $P(f|G, D) = f(G)$ and

$$\hat{P}(f|D) = \sum_{G \in \mathcal{G}} f(G)\hat{P}(G|D). \quad (9)$$

If we are interested in predicting the posteriors of future observations. Let $D^T$ be a set of new data examples sampled independently of $D$ (i.e., $D$ and $D^T$ are independent and identically distributed). Then

$$\hat{P}(D^T|D) = \sum_{G \in \mathcal{G}} P(D^T|G, D)\hat{P}(G|D), \quad (10)$$

and

$$\ln P(D^T|G, D) = \ln P(D^T, D|G)/P(D|G) \quad (11)$$
$$= score(G : D^T, D) - score(G : D). \quad (12)$$

## 3 Finding the $k$-best Network Structures

We find the $k$-best structures using the dynamic programming techniques extending the algorithm for finding the optimal Bayesian network structures in [13]. Our algorithm consists of three steps:

1. Compute the local scores for all possible $n2^{n-1}$ $(i, Pa_i)$ pairs.

2. For each variable $i \in V$, find the $k$-best parent sets in parent candidate set $C$ for all $C \subseteq V \setminus \{i\}$.

3. Find the $k$-best networks.

Step 1 is exactly the same as in [13] and we will use their algorithm. Assuming that we have calculated all the local scores, next we describe how to accomplish Steps 2 and 3 using dynamic programming technique.

### 3.1 Finding the $k$-best parent sets

We can find the $k$-best parent sets for a variable $v$ from a candidate set $C$ recursively. The $k$-best parent sets in $C$ for

---

**Algorithm 1** Finding the $k$-best parent sets for variable $v$ from a candidate set $C$

**Input:**
$score_v(C)$: local scores
$bestParents_v[S]$: priority queues of the $k$-best parent sets for variable $v$ from candidate set $S$ for all $S \subseteq C$ with $|S| = |C| - 1$

**Output:**
$bestParents_v[C]$: a priority queue of the $k$-best parents of $v$ from the candidate set $C$

Initialize $bestParents_v[C]$
**for all** $S \subseteq C$ such that $|S| = |C| - 1$ **do** {
$\quad bestParents_v[C] \leftarrow Merge(bestParents_v[C],$
$\quad\quad bestParents_v[S])$
}
Insert $C$ into $bestParents_v[C]$ if $score_v(C)$ is larger than the minimum score in $bestParents_v[C]$

---

$v$ are the $k$-best sets among the whole candidate set $C$ itself, and the $k$-best parent sets for $v$ from each of the smaller candidate sets $\{C \setminus \{c\}|c \in C\}$. Therefore, to compute the $k$-best parent sets for $v$ for every candidate set $C \subseteq V \setminus \{v\}$, we start with sets of size $|C| = 1$, then consider sets of $|C| = 2$, and so on, until the set $C = V \setminus \{v\}$.

The skeleton algorithm for finding the $k$-best parent sets for $v$ from a candidate set $C$ is given in Algorithm 1, where we use $bestParents_v[S]$ to denote the $k$ best parent sets for variable $v$ from candidate set $S$ stored in a priority queue, and the operation $Merge(.,.)$ outputs a priority queue of the $k$ best parents given the two input priority queues of $k$ elements. Assuming that the merge operation takes time $T(k)$, finding the $k$-best parent sets for $v$ from a candidate set $C$ takes time $O(T(k) * |C|)$, and computing for all $C \subseteq V \setminus \{v\}$ takes time $\sum_{|C|=1}^{n-1} T(k) * |C| * \binom{n-1}{|C|} = O(T(k)(n-1)2^{n-2})$.

### 3.2 Finding the $k$-best network structures

Having calculated the $k$-best parent sets for each variable $v$ from any set $C$, finding the $k$-best network structures over a variable set $W$ can be done recursively. We will exploit the fact that every DAG has a *sink*, a node that has no outgoing edges. First for each variable $s \in W$, we can find the $k$-best networks over $W$ with $s$ as a sink. Then the $k$-best networks over $W$ are the $k$-best networks among {the $k$-best networks over $W$ with $s$ as a sink : $s \in W$}.

The $k$-best networks over $W$ with $s$ as a sink can be identified by looking at the $k$-best parent sets for $s$ from the set $W \setminus \{s\}$ and the $k$-best networks over $W \setminus \{s\}$.[1] More formally, let $bestParents_s[C][i]$ denote the $i$th best parent set for variable $s$ in the candidate parent set $C$. Let

---
[1] This is because for any networks not constructed out of these parents sets for $s$ and networks over $W \setminus \{s\}$ we can always produce $k$ better networks.

**Algorithm 2** Finding the $k$-best network structures over set $W$

**Input:**
$bestParents_i[S]$: priority queues of $k$-best parent sets for each variable $i \in V$ from any candidate set $S \subseteq V - \{i\}$
$bestNets[S]$: priority queues of $k$-best network structures over all $S \subseteq W$ with $|S| = |W| - 1$
**Output:**
$bestNets[W]$: a priority queue of $k$-best networks over $W$

**for all** $s \in W$ **do** {
  **do** a best-first graph search over the space $\{(i, j)\}$ **until** $value(i, j) < score(bestNets[W][k])$ {
    Construct a BN $G$ from the network $bestNets[W \setminus \{s\}][j]$ and setting the set $bestParents_s[W \setminus \{s\}][i]$ as the parents of $s$
    Insert the network $G$ into $bestNets[W]$ if $G$ is not in the queue yet
  }
}

$bestNets[W][j]$ denote the $j$th best network over $W$. Define the function $value(i, j)$ as

$$value(i,j) = score_s(bestParents_s[W \setminus \{s\}][i]) \\ + score(bestNets[W \setminus \{s\}][j]). \quad (13)$$

Then the $k$-best networks over $W$ with $s$ as a sink can be identified by finding the $k$-best scores among

$$\{value(i,j) : i = 1, \ldots, k, \quad j = 1, \ldots, k.\}. \quad (14)$$

This can be done by using a standard best-first graph search algorithm over a search space $\{(i, j) : i = 1, \ldots, k, \; j = 1, \ldots, k\}$ with root node $(1, 1)$, children of $(i, j)$ being $(i+1, j)$ and $(i, j+1)$, and the value of each node given by $value(i, j)$.

The skeleton algorithm for finding the $k$-best network structures over a set $W$ is given in Algorithm 2. Let the time spent on the best-first search be $T'(k)$. In the worst case all $k^2$ nodes may need to be visited. The complexity of finding the $k$-best network structures is $\sum_{|W|=1}^{n} T'(k) * |W| * \binom{n}{|W|} = O(T'(k)n2^{n-1})$.

## 4 Experiments

We used BDe score for $score_i(Pa_i : D)$ with a uniform structure prior $P(G)$ and equivalent sample size 1 [6]. We have implemented our algorithm in C++ language and all the experiments on our method were run under Linux on an ordinary desktop PC with a 3.0GHz Intel Pentium processor and 2.0GB memory.

We tested our algorithm on several data sets from the UCI Machine Learning Repository [1]: Iris, Nursery, Tic-Tac-Toe, Zoo and Letter. We also tested our algorithm on synthetic data sets of various sample sizes from a gold-standard 15-variable Bayesian network with known structure and parameters. All the data sets contain discrete variables (or are discretized) and have no missing values.

### 4.1 Performance evaluation

For each data set, we learned the $k$-best networks for certain $k$, then we can estimate the posterior probabilities $\hat{P}(h|D)$ of any hypotheses using Eqs. (7) and (8). To get an idea on how close the estimation is to the true posteriors we used the algorithm in [16] to compute the exact $P(D)$. We can then evaluate the quality of the posterior estimation as follows. Define the following quantity:

$$\Delta \equiv \frac{\sum_{G \in \mathcal{G}} P(G, D)}{P(D)} = \sum_{G \in \mathcal{G}} P(G|D). \quad (15)$$

$\Delta$ represents the cumulative true probability mass of the graphs in $\mathcal{G}$. From Eqs. (1) and (7) we obtain

$$\frac{P(G|D)}{\hat{P}(G|D)} = \Delta. \quad (16)$$

Note that $\Delta \leq 1$ and the larger value of $\Delta$ means the closer estimation $\hat{P}(G|D)$ to exact $P(G|D)$. In general we have the following results on the quality of estimation $\hat{P}(h|D)$.

**Proposition 1**

$$-(1-\Delta)\hat{P}(h|D) \leq P(h|D) - \hat{P}(h|D) \leq (1-\Delta)(1 - \hat{P}(h|D)), \quad (17)$$

*or equivalently*

$$\Delta \hat{P}(h|D) \leq P(h|D) \leq \Delta \hat{P}(h|D) + 1 - \Delta. \quad (18)$$

The proof is given in the Appendix.

In practice, in the cases we do not have a large amount of data, $\Delta$ may be much smaller than 1 and the bounds in proposition 1 could be too loose to be useful. Therefore, we introduce another measure for the quality of the posterior estimation, the relative ratio of the posterior probability of the MAP network $G_{map}$ over the posterior of the worst network $G_{min}$ of the $k$ best networks (the $k$-th best network):

$$\lambda \equiv \frac{\hat{P}(G_{map}|D)}{\hat{P}(G_{min}|D)} = \frac{P(G_{map}|D)}{P(G_{min}|D)} \quad (19)$$

It has been argued in [10] that we should make predictions using a set of the best models discarding those models that predict the data far less well even though the very many models with small posterior probabilities may contribute substantially to the sum (such that $\Delta$ is much smaller than 1). A cutoff value of $\lambda = 20$ is suggested in [10] by analogy with the 0.05 cutoff for P-values.

Table 1: Experimental Results

| Name | $n$ | $m$ | $T_l$ | $k$ | $T_t$ | $\Delta$ | $\lambda$ |
|---|---|---|---|---|---|---|---|
| | | | Statistics (time in sec.) | | | | |
| Iris | 5 | 150 | 0.3 | 900 | 1.3 | 1.000 | 2.08e+6 |
| Nursery | 9 | 12960 | 0.4 | 100 | 2.9 | 1.000 | 2.07e+16 |
| Tic-Tac-Toe | 10 | 958 | 0.3 | 1000 | 261 | 0.759 | 2.17e+4 |
| Zoo | 17 | 101 | 9 | 1 | 22 | 1.31e-10 | 1 |
| | | | | 10 | 131 | 1.24e-09 | 1.089 |
| | | | | 100 | 5945 | 1.013e-08 | 1.516 |
| Letter | 17 | 20000 | 277 | 1 | 290 | 0.00159 | 1 |
| | | | | 10 | 380 | 0.0159 | 1 |
| | | | | 100 | 5976 | 0.156 | 1.022 |
| Synthetic | 15 | 200 | 4 | 1 | 7 | 1.55e-07 | 1 |
| | | | | 10 | 30 | 1.55e-06 | 1 |
| | | | | 100 | 889 | 1.27e-05 | 1.698 |
| Synthetic | 15 | 1000 | 11 | 1 | 14 | 2.54e-06 | 1 |
| | | | | 10 | 36 | 2.37e-05 | 1.203 |
| | | | | 100 | 884 | 7.00e-05 | 12.49 |
| Synthetic | 15 | 3000 | 17 | 1 | 20 | 4.77e-07 | 1 |
| | | | | 10 | 41 | 3.37e-06 | 1.319 |
| | | | | 100 | 886 | 3.77e-06 | 1282 |
| Synthetic | 15 | 5000 | 21 | 1 | 24 | 2.59e-07 | 1 |
| | | | | 10 | 45 | 1.80e-06 | 4.30 |
| | | | | 100 | 874 | 2.94e-06 | 2406 |

Table 2: The difference across 2 best equivalence classes

| Name | No. of diff. edges | $\lambda$ |
|---|---|---|
| Iris | 2 | 2.30 |
| Nursery | 1 | 15.32 |
| Tic-Tac-Toe | 4 | 1.00 |
| Zoo | 1 | 1.03 |
| Letter | 1 | 1.02 |
| Synthetic ($m = 200$) | 1 | 1.01 |
| Synthetic ($m = 1000$) | 1 | 1.03 |
| Synthetic ($m = 5000$) | 1 | 4.30 |

### 4.2 Experimental results on the $k$-best networks

We tested our algorithm on several data sets and the experimental results are reported in Table 1, which lists for each data set the number of variables $n$, the number of instances (sample size) $m$, the value $k$, the time $T_l$ for computing local scores (the time depends on $m$ but not $k$), the total running time $T_t$ for finding the $k$-best networks, and the quality measure $\Delta$ and $\lambda$.

As expected the values of $\Delta$ and $\lambda$ (as a measure of estimation quality) increases as $k$ increases. We see that $\Delta$ value is often too small for Proposition 1 to be useful, however Proposition 1 can indeed provide guarantee on the quality of approximation in nontrivial cases like Nursey and Tic-Tac-Toe data sets which are still too large for exhaustive enumeration of all possible networks. Based on the $\lambda$ value, the best 100 networks are not enough to get reliable estimation for Zoo and letter data sets. For the synthetic data set, $\lambda$ increases as the sample size $m$ increases. Based on the $\lambda$ value, the best 100 networks should give reliable estimation for $m = 3000$ and $m = 5000$.

The exact posterior probabilities of $k$-best networks for each data set are shown in Figures 1 and 2. We see that there could be multiple networks having the same posterior probability. For example, the number of best networks sharing the largest posterior probability is as follows: 2 for Nursery case, 76 for Tic-Tac-Toe case, 12 for Letter case, 26 for synthetic data with $m = 200$, and 6 for synthetic data with $m = 5000$. This is mainly due to that BDe scoring criterion has the likelihood-equivalence property, i.e., it assigns the same score to the Bayesian networks within the same independence equivalence class. However, we have found that it is also possible that the networks across the different equivalence classes have the same posterior probability. Take Tic-Tac-Toe case for example, we have found that the 76 best networks actually belong to multiple equivalence classes which have different skeletons. This exceptional case shows that one can not always assert that the networks having the same posterior probability must be within the same equivalence class.

In Table 2 we show the difference between the best equivalence class and the second best equivalence class for each case. The second column lists the number of different undirected edges between these 2 best equivalence classes and the third column shows $\lambda$ (the relative ratio of their posterior probabilities). The result shows that their difference is typically small for each case.

### 4.3 Structural discovery

In order to evaluate the ability of the $k$-best networks method in structural discovery, we tested on the synthetic data set from the gold-standard 15-variable Bayesian network. We computed edge feature between each pair of variables by Eq. (9) under the different values of $k \in \{1, 10, 100\}$. For the comparison, we also computed the exact posterior probability of each edge by the method of full model averaging in [16]. Since we have the true gold-standard network, we could compute all the corresponding ROC curves. The results are shown in Figures 3 and 4.

The figures indicate the usefulness of $k$-best method in structural discovery. We observe that the area under ROC (AUC) is a non-decreasing function of $k$. Even a small increase of $k$ from 1 to 10 will lead to a non-negligible improvement in the corresponding ROC, even though $\Delta$ is tiny (such as $2.37e-05$) and $\lambda$ is small (such as 1.203). For the data set with $m = 5,000$, the performance of Top 100 is almost the same as that of full model averaging method when $\lambda$ is big (2,406), regardless of the fact that $\Delta$ is still tiny ($2.94e - 06$).

The comparison with MCMC approach also demonstrates the usefulness of our method. In this paper we compared our method with the hybrid method (DP+MCMC) proposed by Eaton and Murphy [3], which was shown to have the statistically significant improvement in structural discovery over other MCMC methods [5, 11]. For DP+MCMC, we set the pure global proposal (with local proposal choice $\beta = 0$) since such a setting was reported to have the best performance (with the largest mean and the smallest variance of AUC) for edge discovery in their experimental results. The tool BDAGL provided by the authors in [3] was used for the experiments on DP+MCMC. We ran totally $120,000$ iterations and discarded the first $100,000$ iterations as burn-in period. Then we set the thinning parameter as 2 to get final $10,000$ samples (networks). Finally the posterior probability of each edge was computed based on the model averaging among these $10,000$ networks and the corresponding ROC was drawn.

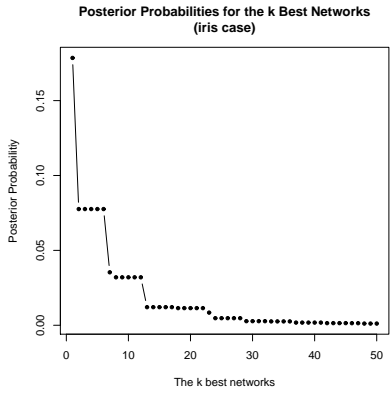
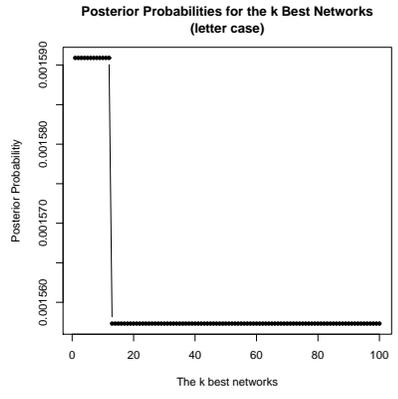
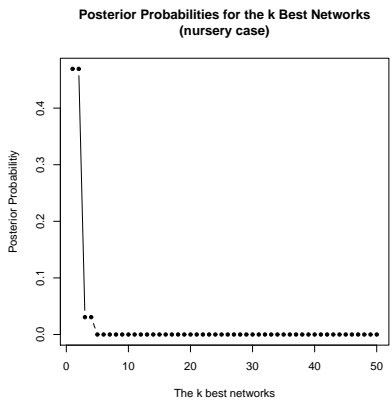
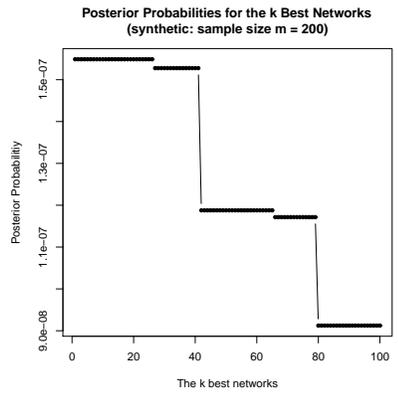
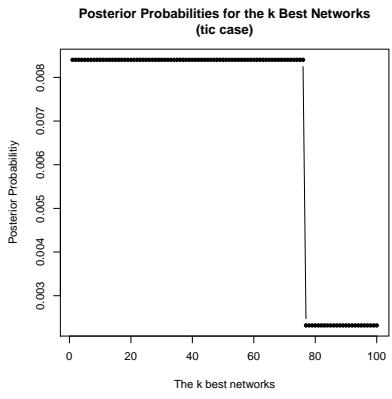
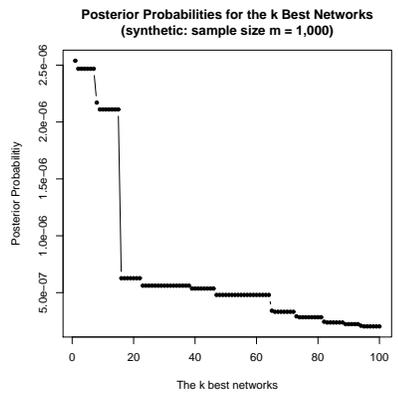
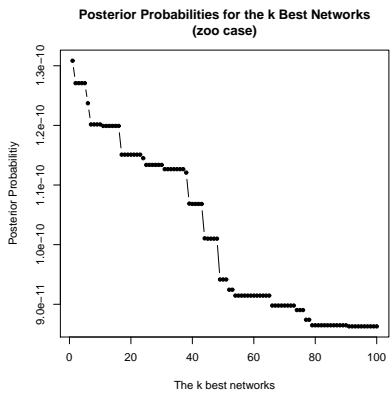
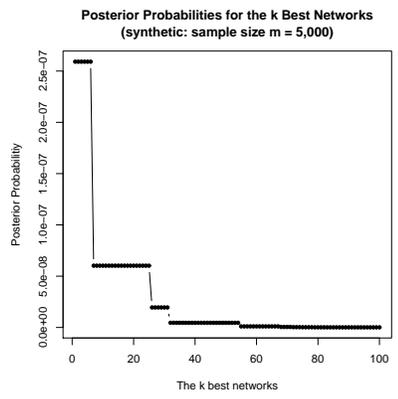

Figure 1: The Exact Posterior Probabilities of the $k$-best Networks

Figure 2: The Exact Posterior Probabilities of the $k$-best Networks (Continued)

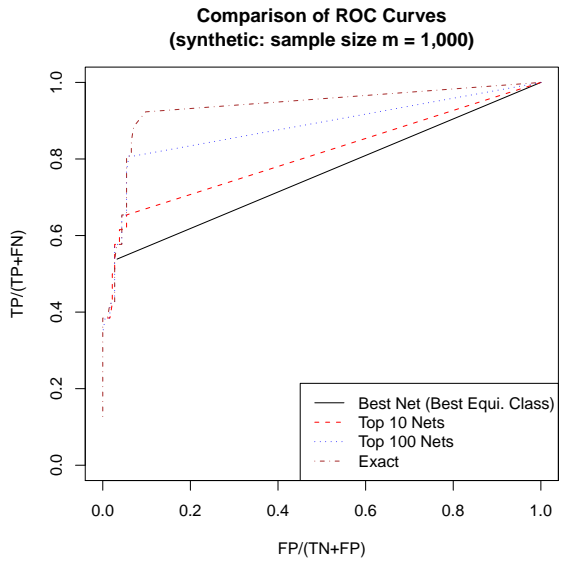
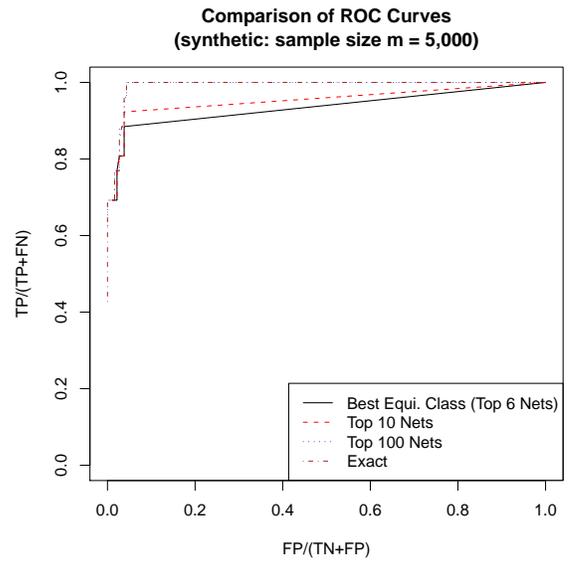
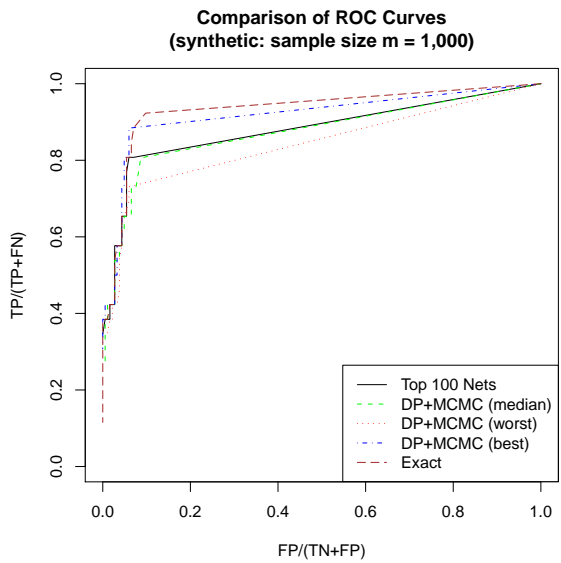
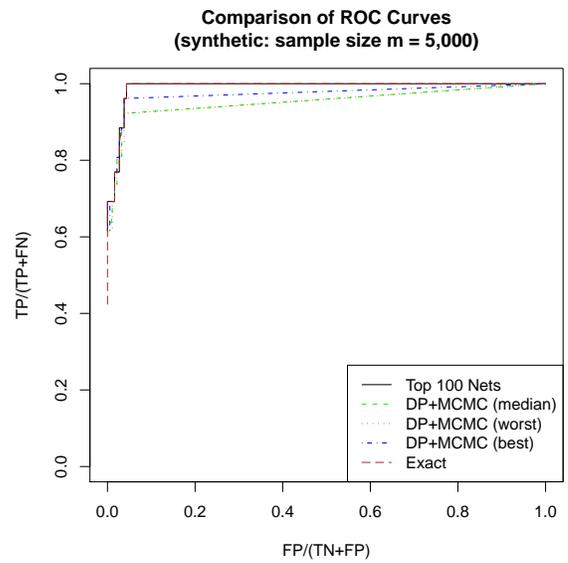

Figure 3: Comparison of ROC Curves ($m = 1,000$)

Figure 4: Comparison of ROC Curves ($m = 5,000$)

Because of the randomness nature of MCMC, we repeated the above process 10 times for each data set.[2] The best, worst and median of these 10 ROCs were shown in the figures, compared with the ROC from Top 100 (i.e. the 100 best). In the case of $m = 1,000$, the figure shows that the performance of DP+MCMC still has a non-negligible variability and the performance of Top 100 is no worse than the performance of the median of DP+MCMC. In the case of $m = 5,000$, the variability of the performance of DP+MCMC decreases. However, even the best of DP+MCMC could not outperform Top 100. The good performance of our model averaging over only 100 networks is not surprising: the 100 networks used here are the top 100 networks and have the relative importance than all the other networks in the model average process. Figure 2 has clearly demonstrated such a relative importance of these top 100 networks.

## 5 Conclusion

We develop an algorithm for finding the $k$-best Bayesian network structures. We present empirical results on the structural discovery by Bayesian model averaging over the $k$ best Bayesian networks. One nice feature of the method is that we can monotonically improve the estimation accuracy by spending more time to compute for larger $k$. Another interesting feature shown by our experiments is that we may evaluate the quality of the estimation based on the value of $\lambda$. The relation between the estimation quality and $\lambda$ is worth more substantial study in the future.

As the experimental results show, there are many equivalent networks in the set of best networks. It is desirable if we can directly find the $k$ best equivalence classes. However, the proposed algorithm does not search in the equivalence class space. The algorithm will find the top $k$ individual networks regardless of the existence of equivalent networks. It seems that the dynamic programming idea cannot be naturally generalized to the equivalence class space (at least we were not able to achieve this). How to directly find the $k$ best equivalence classes is a research direction that is worth to pursue.

---

[2]The DP step (including marginal likelihood computation) took 221 seconds and MCMC iterations took the mean of 134 seconds in the case of $m = 1,000$. The most part of BDAGL was written in Matlab and we ran BDAGL under Windows XP on an ordinary laptop with 1.60GHz Intel Pentium processor and 1.5GB memory. Due to the different hardware, platforms and programming languages used, the time statistics of DP+MCMC can not be directly compared with the ones of our method.

## Appendix: Proof of Proposition 1

$$\begin{aligned}
&P(h|D) - \hat{P}(h|D) \\
&= \sum_G P(h|G,D)P(G|D) - \sum_{G \in \mathcal{G}} P(h|G,D)\hat{P}(G|D) \\
&= \sum_{G \in \mathcal{G}} P(h|G,D)[P(G|D) - \hat{P}(G|D)] \\
&\quad + \sum_{G \notin \mathcal{G}} P(h|G,D)P(G|D) \quad (20) \\
&= \sum_{G \in \mathcal{G}} P(h|G,D)\hat{P}(G|D)(\Delta - 1) \\
&\quad + \sum_{G \notin \mathcal{G}} P(h|G,D)P(G|D) \quad (21) \\
&= -(1-\Delta)\hat{P}(h|D) + \sum_{G \notin \mathcal{G}} P(h|G,D)P(G|D) \quad (22)
\end{aligned}$$

Since the second term in Eq. (22) is no less than zero we have proved

$$P(h|D) - \hat{P}(h|D) \geq -(1-\Delta)\hat{P}(h|D). \quad (23)$$

Since $P(h|G,D) \leq 1$, from Eq. (22) we have

$$\begin{aligned}
&P(h|D) - \hat{P}(h|D) \\
&\leq -(1-\Delta)\hat{P}(h|D) + \sum_{G \notin \mathcal{G}} P(G|D) \\
&= -(1-\Delta)\hat{P}(h|D) + 1 - \Delta \\
&= (1-\Delta)(1-\hat{P}(h|D)). \quad (24)
\end{aligned}$$